# DRIFT: Data Reduction via Informative Feature Transformation- Generalization Begins Before Deep Learning starts

Author: Ben Keslaki

Email: benkeslaki@gmail.com

**Abstract**

Modern deep learning architectures excel at optimization, but only after the data has entered the network. The true bottleneck lies in preparing the right input: minimal, salient, and structured in a way that reflects the essential patterns of the data. We propose DRIFT (Data Reduction via Informative Feature Transformation), a novel preprocessing technique inspired by vibrational analysis in physical systems, to identify and extract the most resonant modes of input data prior to training. Unlike traditional models that attempt to learn amidst both signal and noise, DRIFT mimics physics perception by emphasizing informative features while discarding irrelevant elements. The result is a more compact and interpretable representation that enhances training stability and generalization performance. In DRIFT, images are projected onto a low-dimensional basis formed by spatial vibration mode shapes of plates, offering a physically grounded feature set. This enables neural networks to operate with drastically fewer input dimensions (~ 50 features on MNIST and less than 100 on CIFAR100) while achieving competitive classification accuracy. Extensive experiments across MNIST and CIFAR100 demonstrate DRIFT's superiority over standard pixel-based models and PCA in terms of training stability, resistance to overfitting, and generalization robustness. Notably, DRIFT displays minimal sensitivity to changes in batch size, network architecture, and image resolution, further establishing it as a resilient and efficient data representation strategy. This work shifts the focus from architecture engineering to input curation and underscores the power of physics-driven data transformations in advancing deep learning performance.

Keywords: Feature representation, Neural network, Dimensionality reduction, Generalization gap, Training Stability

Codes available at link: https://github.com/keslaki/DRIFT

## Introduction

A model that performs well during training may not necessarily achieve similar success on unseen data. This discrepancy, commonly referred to as the generalization gap, is a central concern in deep learning. The generalization gap reflects the difference between a model's training and test performance, and a large gap typically signals overfitting, where the model memorizes training data but fails to generalize. Given that the ultimate goal of a deep learning model is to produce reliable predictions on previously unseen data, increasing attention is being directed toward understanding the causes of poor generalization and developing algorithms that mitigate it.

A growing body of literature explores various facets of this problem. For instance, Keskar et al. [1] noted that larger batch sizes often reduce generalization performance, and they proposed numerical strategies to investigate this phenomenon. Hoffer et al. [2] argued that the observed gap stems more from fewer update steps rather than batch size itself and demonstrated that appropriate training strategies can fully mitigate the gap. Schmidt et al. [3] challenged the conventional wisdom that more training data always improves robustness, showing experimentally (even with linear regression) that increasing data size can widen the generalization gap in adversarial settings. Other works [4–6] have investigated the interplay between generalization, optimization, and calibration, noting that strategies like data augmentation or smaller models can help. However, augmentation can also lead to overfitting when models learn spurious correlations rather than fundamental patterns. Similarly, research in reinforcement learning [7–9] has shown that environmental factors and initial state distributions heavily influence generalization. Several studies [10–12] have debunked the assumption that model complexity, as measured by parameter count, directly correlates with overfitting. Zhang et al. [13] highlighted that deep networks can memorize even random labels, raising questions about what truly enables generalization. Instead, they suggest that deep networks generalize well despite being overparameterized. Recent theoretical advances, such as the Coherent Gradients theory [14], propose that gradient alignment during training can lead to stable, generalizable solutions, offering a dynamic view that contrasts with traditional static complexity measures.

This study initiates our exploration of physics-inspired data preprocessing for deep learning. Initial experiments with feedforward neural networks on datasets such as MNIST and CIFAR100 demonstrate promising results, achieving effective generalization with significantly fewer features. Future work will extend DRIFT to convolutional neural networks (CNNs) to enhance hierarchical feature extraction and focus on informative spatial patterns, potentially further reducing redundancy.

**Modeling**

In this approach, we model a 2D image as a physical structure capable of vibrating, much like a real-world object. Just as physical objects exhibit natural modes of vibration depending on their shape and support conditions, we treat an image as a thin rectangular plate (in our case, a 28×28 pixel grid like MNIST digits) that can vibrate with distinct mode shapes which shows a vibrating pattern with the vibration amplitude being the pixel values at x and y coordinates as similarly shown in Fig 1.

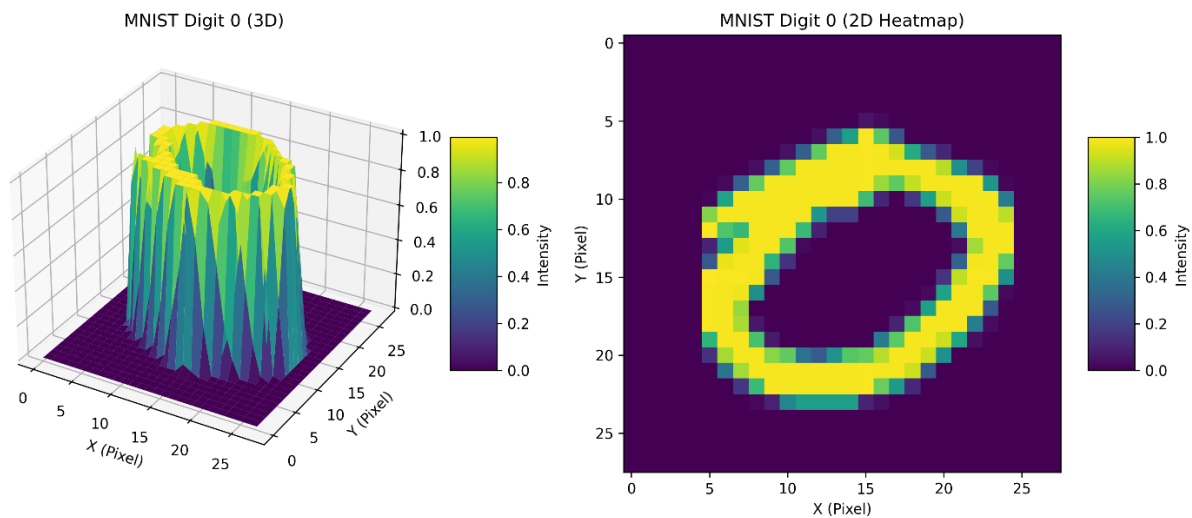

Figure 1. Sample MNIST digit image, where pixel intensities are analogous to the vibration amplitudes of a simply supported plate, illustrating the correspondence between image data and vibrational mode shapes.

A simply supported plate's vibrational mode shapes, described by $sin\left(\frac{n\pi x}{L_x}\right) \times sin\left(\frac{m\pi y}{L_y}\right)$ [15], where $n$ and $m$ are positive integers, $x$ and $y$ are spatial coordinates, and $L_x$ and $L_y$ are the plate's dimensions (28×28 for MNIST), provide a physics-inspired framework for image analysis shown in figure 2 as 2D and 3D representation. Each pixel in an image can be interpreted as the amplitude of vibration at position ($x,y$), analogous to a vibrational mode or a combination of modes. In a 28×28 grid, the system has 784 theoretical degrees of freedom, each corresponding to a unique spatial deformation pattern. By visualizing these mode shapes for different ($n,m$) pairs, we capture the foundational behaviors of the plate, with the first few modes being the most dominant and influential due to their ease of excitation and significant contribution to the system's dynamics [16]. To apply this framework to image recognition, we compute the cosine similarity between an image and a set of predefined vibrational mode shapes, generating a feature vector with N being the number of modes considered. This transformation reduces the original 784-pixel image to an N×1 feature vector, significantly lowering input dimensionality. When used in a deep learning model, this approach yields faster and more stable training convergence, improved generalization with a reduced training-testing performance gap, and lower computational complexity, demonstrating the effectiveness of focusing on lower-order modes for feature extraction.

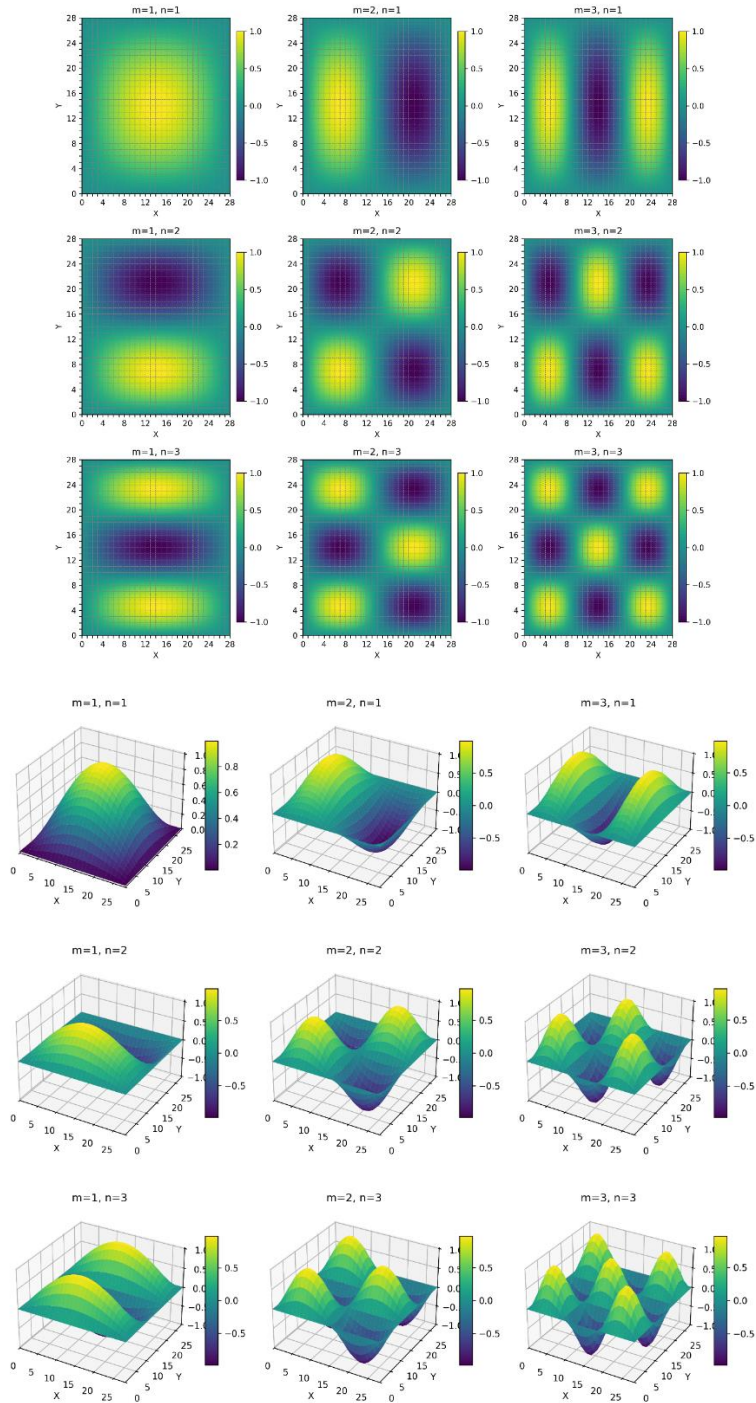

Figure 2. Nine sample of two-dimensional and three-dimensional visualizations of vibrational mode shapes used for cosine similarity computations in feature extraction, illustrating their spatial patterns for image analysis FOR A GRID OF 28×28.

**Experiments**

**MNIST:**

In this section, we analyze the comparative performance of three different feature representations, DRIFT, PCA, and the Full model (standard NN without dimensionality reduction), as illustrated in Figure 3. All models share a consistent NN architecture comprising three hidden layers [64, 128, 64] and are trained using a batch size of 32. The primary variables in our experiments are the number of retained features (modes), which are set to 20, 30, and 50 for subfigures (a–f). Each model is evaluated on training and test accuracy as well as loss to provide a holistic view of performance. For the lowest-dimensional setting with 20 modes, the DRIFT-based model demonstrates a notable advantage. Despite operating with significantly fewer features than the Full model (20 vs. 784), DRIFT achieves accuracy levels that are comparable to PCA, as seen in Figure 3a. More importantly, DRIFT exhibits much greater stability during training, with smoother convergence and lower variance in accuracy. In terms of loss (Figure 3b), DRIFT consistently outperforms PCA, showing lower values on the test datasets. This indicates not only better optimization behavior but also superior generalization. Compared to the Full model, DRIFT incurs a slight drop in accuracy, which is expected due to the reduced input space, but this trade-off is offset by the model's compactness and lower test loss. These results suggest that DRIFT is highly effective at extracting informative features while avoiding overfitting. As the number of modes increases to 30, both DRIFT and PCA show improvements in accuracy, approaching the performance of the Full model. However, DRIFT continues to maintain a smoother and more stable training trajectory. The loss curves further support DRIFT's advantage, with reduced fluctuations and better alignment between training and test loss. This indicates improved generalization and reduced overfitting. The results suggest that moderate feature dimensionality allows DRIFT to balance expressive power and model stability effectively. In the case of 50 modes, both DRIFT and PCA models come very close to matching the Full model in terms of accuracy. However, we observe an increase in oscillatory behavior in the loss curves, especially in the DRIFT model. This suggests that while adding more features increases representational capacity, it also introduces the risk of model confusion or overfitting, as indicated by the larger test loss. The early onset of oscillations in training loss with higher mode counts highlights a potential drawback of high-dimensional embeddings: beyond a certain point, the added complexity may degrade rather than improve model performance.

Overall, the comparative analysis underscores the effectiveness of DRIFT as a feature reduction technique. Across all configurations, DRIFT outperforms PCA in terms of stability, convergence, and loss, while remaining competitive in accuracy, even when using a fraction of the input features. Although increasing the number of modes does improve accuracy, it may also lead to training instability. Thus, DRIFT with a carefully chosen mode count provides an excellent trade-off between performance, stability, and computational efficiency, making it a promising approach for compact and interpretable neural network design.

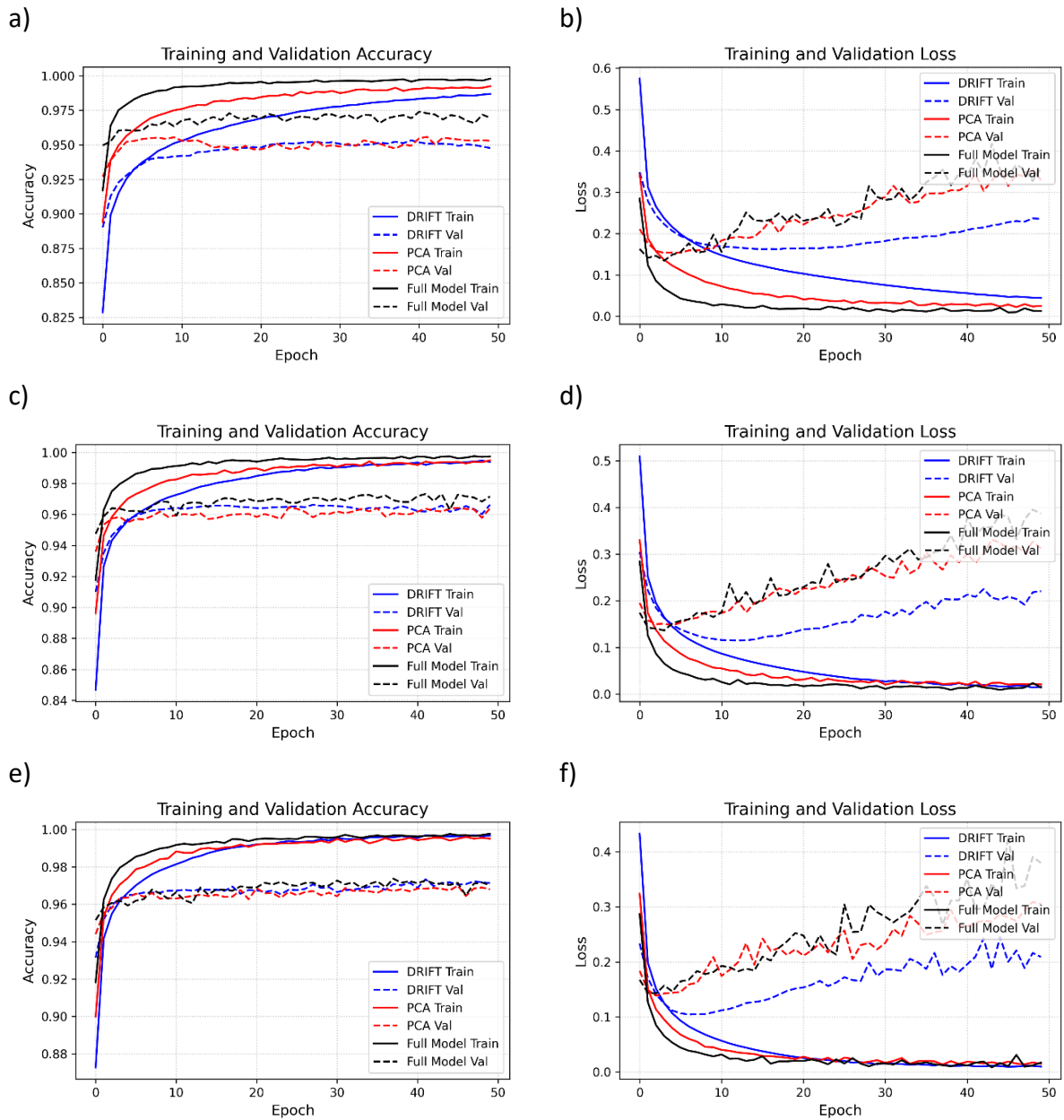

Figure 3. Comparison of DRIFT, PCA, and full models across varying mode counts, with a batch size of 32 and hidden layers configured as [64, 128, 64]. Subfigures depict performance for mode counts: (a, b) 20 modes, (c, d) 30 modes, and (e, f) 50 modes.

In Figure 4, we investigate the interaction between batch size and model performance for a fixed number of modes (30). This analysis is performed across three batch sizes: 2 (Figures 4a–b), 32 (Figures 4c–d), and 256 (Figures 4e–f), with a focus on the behavior of dimensionality-reduction-based NN models: DRIFT, PCA, and the Full model (baseline NN using all original input features). It is well-established that smaller batch sizes typically result in noisier gradient updates, leading to higher variance in the loss and accuracy, particularly on the test set. This is clearly observed in the Full model, where the use of a batch size of 2 results in a notably high training loss and significant instability in convergence. As the batch size increases to 32 and then to 256, the loss for the Full model decreases and convergence improves, highlighting its dependence on batch size for stable learning. In contrast, the DRIFT model demonstrates remarkable stability across all batch sizes. Even under the most challenging condition (batch size = 2), DRIFT maintains smooth convergence and competitive test accuracy. Not only does it outperform PCA in terms of both accuracy and loss, but it also resists the oscillatory behavior typically induced by small batch sizes. This suggests that DRIFT's reduced feature space effectively captures the core structural relationships within the data, thereby shielding the model from noise-induced instability. Interestingly, changing the batch size has minimal impact on DRIFT's performance, which strongly supports the claim that DRIFT is less sensitive to optimization hyperparameters compared to PCA and the Full model. This stability suggests that DRIFT selects and retains robust, high-information components of the dataset, preserving the essential dynamics of the learning task while filtering out irrelevant noise. Such a property is particularly valuable in real-world scenarios where training conditions may vary, or computational resources may be limited. From a broader perspective, this experiment supports the notion that pre-training feature reduction plays a critical role in model robustness. By distilling the input to its most relevant features prior to training, DRIFT enables models to generalize better and train more reliably. This aligns with our overarching hypothesis: the quality and relevance of the input features are foundational to model success, and a well-constructed feature space can reduce dependency on both the architecture and the training regimen.

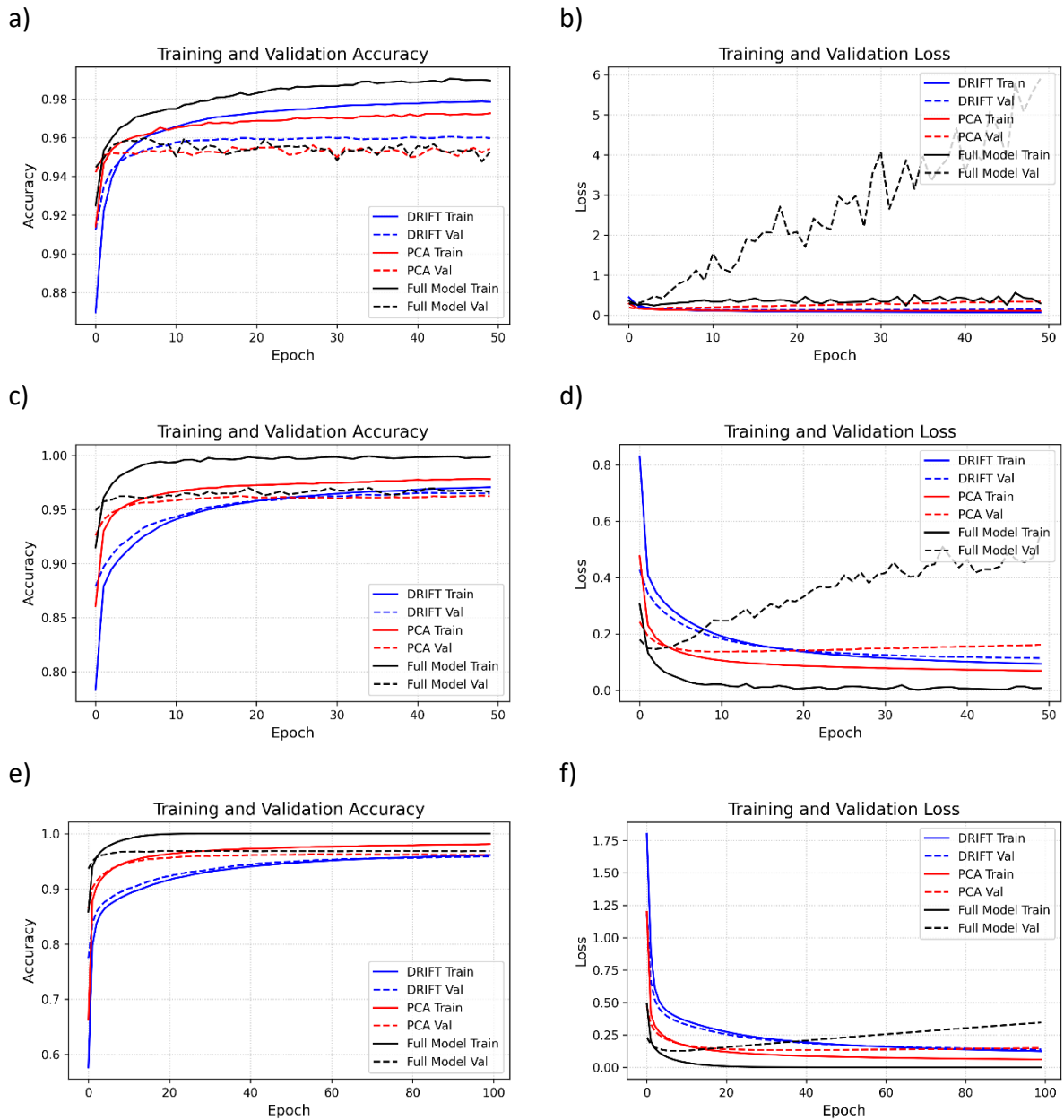

Figure 4. Impact of batch size on model stability and performance, with a fixed mode count of 30 and a single hidden layer of 64 units. Subfigures depict results for batch sizes: (a, b) 2, (c, d) 32, and (e, f) 256.

**CIFAR100:**

In Figure 5, we investigate the impact of varying the number of modes on the performance of three models, DRIFT, PCA, and the Full model, when applied to the CIFAR100 dataset. For this experiment, we maintain the original input shape of the dataset, which consists of 32×32-pixel arrays across three channels. While input resizing is a relevant factor, its effect will be discussed in the context of subsequent figures. The focus in this figure is solely on the number of retained modes used in the dimensionality reduction process. The experiments explore three different mode counts: 40 (Figures 5a and 5b), 80 (Figures 5c and 5d), and 150 (Figures 5e and 5f). In all three cases, DRIFT consistently outperforms both PCA and the Full model in terms of generalization, stability, and test loss. This superiority highlights DRIFT's capability to extract the most informative features with high efficiency. Notably, the Full model's performance remains unchanged across all mode counts, as it does not involve any feature reduction. However, PCA shows improved training accuracy with an increasing number of modes, while its generalization performance (i.e., test accuracy and loss) deteriorates. This reinforces the well-established notion that increasing the feature count does not guarantee improved model generalization.

Interestingly, the DRIFT model exhibits remarkable stability for all the number of modes. Unlike PCA, which becomes more prone to overfitting with more features, DRIFT maintains low sensitivity to such changes. This suggests that DRIFT not only identifies the most impactful components of the data but also filters out noise and redundancy more effectively than PCA. Its ability to maintain consistent performance across a range of input dimensionalities makes it a robust alternative for feature extraction. It is important to note that the classification accuracies reported in these experiments are relatively modest. This is due to the simplicity of the network architecture employed, a basic feedforward neural network with three hidden layers of sizes [64, 128, 64], and the uniform batch size of 128 used for all experiments. Nonetheless, the relative performance trends are clear and informative: DRIFT yields more stable and generalizable results compared to PCA, even when the number of modes is significantly increased. These findings underscore the importance of quality over quantity in feature selection and suggest that DRIFT offers a more effective and scalable strategy for high-dimensional data analysis.

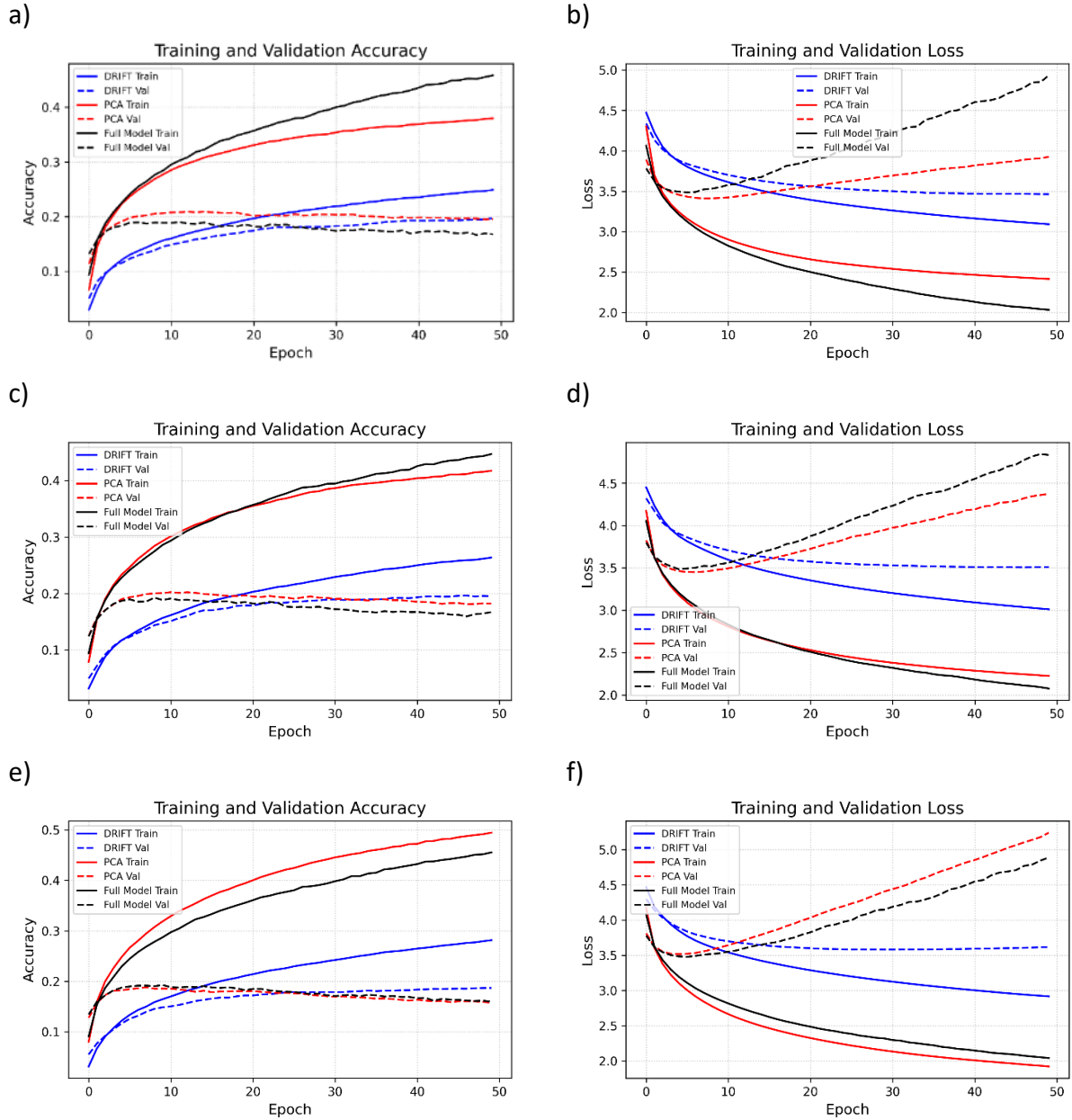

Figure 5. Impact of Mode Count on DRIFT, PCA, and Full Feature Models using the CIFAR100 Dataset. (Input shape: 32×32×3; Batch size: 128; Hidden layers: [64, 128, 64]). Subplots correspond to different mode counts: (a, b) = 40 modes, (c, d) = 80 modes, and (e, f) = 150 modes.

In Figure 6, we present results using an expanded input resolution to assess the sensitivity of different feature reduction approaches to changes in image size. Specifically, the CIFAR100 dataset was resized from its original resolution of 32×32×3 to 80×80×3, while keeping the same network architecture ([64, 128, 64] hidden layers) and a batch size of 128. The number of retained modes was set to 80 for both DRIFT and PCA methods. When comparing these results to the baseline configuration shown in Figure 5 (subplots c and d), which used the standard 32×32×3 input resolution, some difference in sensitivity emerges among the three models. Notably, the DRIFT-based model maintains a consistent level of performance, demonstrating that increasing the input resolution has minimal effect on both its accuracy and generalization behavior. This result highlights DRIFT's low dependency on input dimensionality and reinforces its robustness across varying input configurations. In contrast, PCA exhibits a marked degradation in generalization performance at the higher input resolution. This suggests that PCA may be more sensitive to input redundancy and high-dimensional feature noise, which are amplified by the increased resolution. The Full model, which utilizes the complete set of input features without any dimensionality reduction, shows no substantial change in performance. However, its already limited generalization capability remains largely unimproved, indicating that simply increasing resolution does not enhance its ability to generalize. Overall, the findings from Figure 6 underscore DRIFT's resilience to changes in input size, further positioning it as a stable and effective alternative for feature reduction in deep learning pipelines, especially in scenarios where input dimensionality varies, or computational efficiency is a concern.

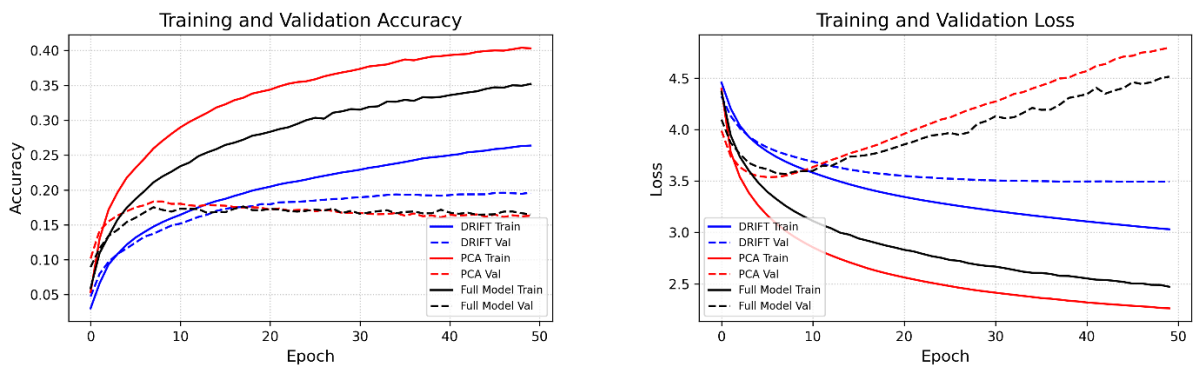

Figure .6. Impact of increased input resolution (80×80×3) on DRIFT, PCA, and Full models for CIFAR100, using a neural network with hidden layers [64, 128, 64] and batch size 128.

Figure 7 investigates the impact of batch size and network density on the training stability and generalization performance of three models: DRIFT, PCA-based, and a standard Full model. For this experiment, we fixed the input size to 32×32×332 (i.e., CIFAR100 image dimensions), used 80 modes as input features, and employed a relatively shallow network with 32 hidden units. Two different batch sizes were tested: 2 and 128, corresponding to subfigures (a, b) and (c, d), respectively. The results clearly demonstrate that larger batch sizes yield smoother convergence behavior. In the case of batch size 128 (Figures 7c and 7d), both PCA and DRIFT models show consistent and stable training trajectories, while the Full model suffers from reduced generalization, shown in the test phase. This observation aligns with known behavior in deep learning, where larger batches often lead to more stable gradient updates and smoother loss landscapes. However, the most interesting insight comes from the comparison between DRIFT and the other models under small batch size conditions (Figures 7a and 7b). Despite the increased noise and instability introduced by the smaller batch size of 2, the DRIFT model maintains stable convergence and generalization, while both PCA and the Full model begin to diverge after some training iterations. This divergence indicates a loss of learning capacity and reliability in these models when subjected to high-variance gradient updates. These findings reinforce the robustness of the DRIFT approach. Unlike PCA and standard neural networks, DRIFT retains its predictive integrity even under more volatile training regimes. Its ability to consistently generalize, even with small batch sizes, underscores the value of selecting meaningful and structured features. This experiment highlights the fact that not only model architecture, but also data representation and preprocessing, play a crucial role in deep learning performance. The resilience of DRIFT to hyperparameter changes such as batch size confirms its suitability as a stable and generalizable feature learning strategy.

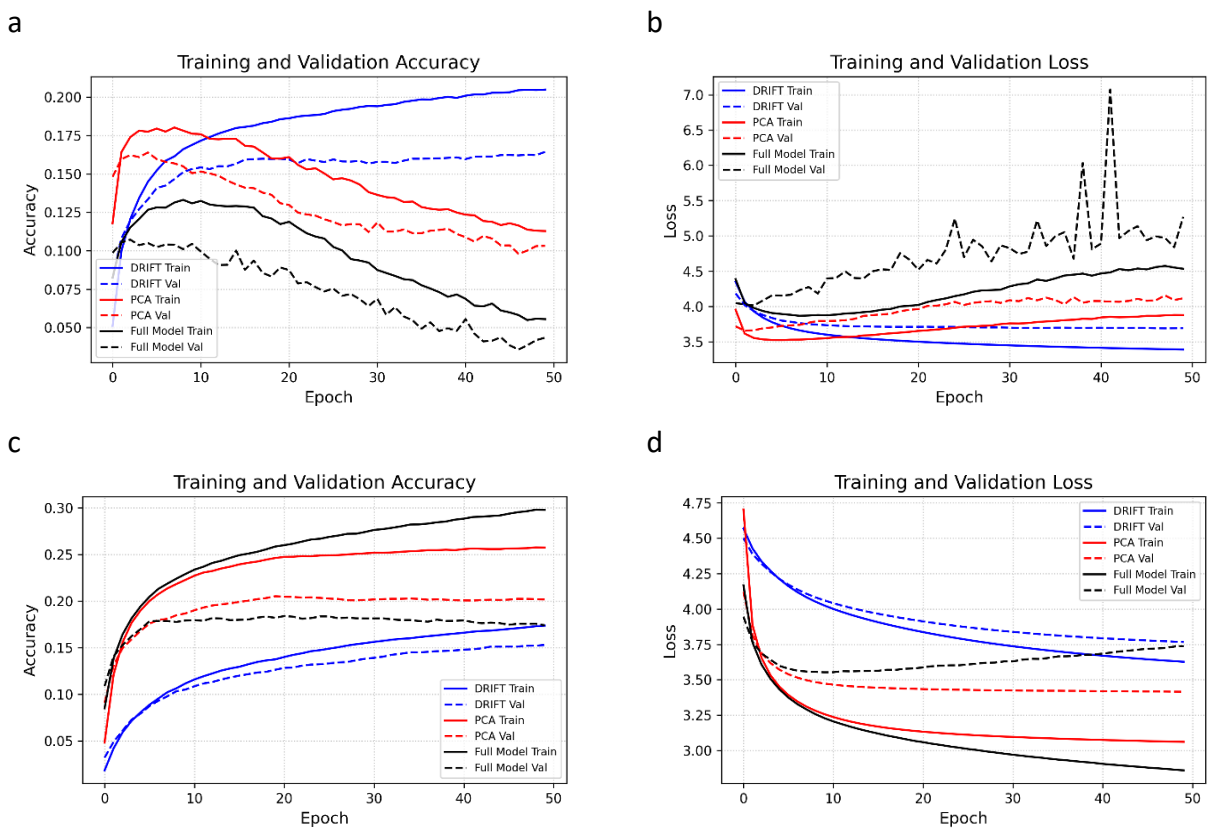

Figure 7. Impact of Batch Size on Model Stability for CIFAR100 using a 32-layer hidden architecture. Results are shown for batch size 2 (a, b) and batch size 128 (c, d).

**Conclusion**

This paper introduced DRIFT, a novel data preprocessing strategy rooted in vibrational mode analysis, which allows neural networks to operate on a compact, high-information representation of the input space. By transforming images into a lower-dimensional modal basis, we significantly reduce model complexity while preserving or even enhancing generalization performance. DRIFT demonstrated superior stability, reduced sensitivity to batch size, and improved training-test alignment compared to both PCA and standard full-input models across diverse datasets including MNIST and CIFAR100. The results affirm our central hypothesis: effective generalization begins with better data representation, not merely with architectural tuning or post-hoc regularization. DRIFT functions as a data distillation pipeline, selectively retaining essential features while discarding redundant or noisy input dimensions. This reduces reliance on large-scale models or extensive training tweaks and positions DRIFT as a general-purpose preprocessing technique for image classification tasks. Importantly, this work marks the initiation of a broader research direction. The current study focuses on feedforward neural networks to establish foundational understanding. In future work, we will extend DRIFT to convolutional neural networks (CNNs) and explore how structured feature extraction can enhance CNN performance, particularly in scenarios where spatial hierarchies and local patterns are paramount. By bridging physical insight with machine learning practice, DRIFT offers a principled approach to more interpretable, efficient, and generalizable deep learning.